\documentclass[letterpaper, 10 pt, conference]{ieeeconf}

\IEEEoverridecommandlockouts                              
\newif\ifcomment
\commenttrue   

\usepackage{amsmath,mathtools} 
\usepackage{amssymb}  
\usepackage{graphicx}
\usepackage{siunitx}
\usepackage{subcaption}
\usepackage{booktabs}
\usepackage{xcolor}
\usepackage{textcomp}
\usepackage{gensymb} 
\usepackage{multirow} 
\usepackage{comment}
\usepackage{verbatim}
\usepackage[skip=0pt, font=footnotesize]{caption}
\usepackage{bm}
\usepackage[noadjust]{cite}
\usepackage{makecell}
\usepackage{tabularx}
\usepackage{pifont}
\usepackage{layouts}
\usepackage{physics}

\usepackage{amsthm}
\usepackage{algorithm}
\usepackage{algpseudocode}
\usepackage{lipsum}
\usepackage{epstopdf}

\usepackage[inkscapeformat=png]{svg}

\usepackage[english]{babel}
\usepackage[hidelinks]{hyperref}

\usepackage[normalem,normalbf]{ulem}
\usepackage{accents}
\newcommand{\ubar}[1]{\underaccent{\bar}{#1}}

\newtheoremstyle{colon}%
{}
{}
{\itshape}
{}
{\bfseries}
{:}
{ }
{}
\theoremstyle{colon}

\newtheorem{problem}{Problem}

\theoremstyle{remark}

\makeatletter
\renewcommand{\@IEEEsectpunct}{ \,}
\makeatother

\makeatletter
\newcommand\footnoteref[1]{\protected@xdef\@thefnmark{\ref{#1}}\@footnotemark}
\makeatother

\makeatletter 
\newcommand\Label[1]{&\refstepcounter{equation}(\theequation)\ltx@label{#1}&}
\makeatother

\definecolor{peru}{rgb}{0.803921568627451, 0.5215686274509804, 0.24705882352941178}
\definecolor{violet}{rgb}{0.9333333333333333, 0.5098039215686274, 0.9333333333333333}
\definecolor{greeN}{rgb}{0.17254901960784313, 0.6274509803921569, 0.17254901960784313}
\definecolor{stage0}{RGB}{187,248,255}
\definecolor{stage1}{RGB}{250,255,187}
\definecolor{stage2}{RGB}{187,255,196}
\definecolor{stage0_dark}{RGB}{0,180,200}
\definecolor{stage1_dark}{RGB}{200,180,0}
\definecolor{stage2_dark}{RGB}{0,200,0}

\definecolor{centerline}{RGB}{51,51,255}
\definecolor{exterior}{RGB}{255,153,51}
\definecolor{interior}{RGB}{0,153,0}
\definecolor{slalom}{RGB}{255,51,51}

\def\anonymous{1} 

\newcommand\ringring[1]{%
  {
   \mathop{\kern0pt #1}\limits^{
     \vbox to-1.85ex{
       \kern-2ex 
       \hbox to 0pt{\hss\normalfont\kern.1em \r{}\kern-.45em \r{}\hss}%
       \vss 
     }
   }
  }
}

\renewcommand{\baselinestretch}{0.98}

\newcolumntype{M}[1]{>{\centering\arraybackslash}m{#1}}

\makeatletter
\def\endthebibliography{%
	\def\@noitemerr{\@latex@warning{Empty `thebibliography' environment}}%
	\endlist
}
\makeatother

\IEEEoverridecommandlockouts
\usepackage{tikz}
\usepackage{textcomp}
\usepackage{hyperref}
\usepackage{lipsum}

\newcommand\copyrighttext{%
	\footnotesize \footnotesize Published in IEEE/RSJ International Conference on Intelligent Robots and Systems (IROS), Abu Dhabi, UAE, 2024.\newline
    \textcopyright 2024 IEEE. Personal use of this material is permitted.
	Permission from IEEE must be obtained for all other uses, in any current or future media, including reprinting/republishing this material for advertising or promotional purposes, creating new collective works, for resale or redistribution to servers or lists, or reuse of any copyrighted component of this work in other works.
 }
\newcommand\copyrightnotice{%
	\begin{tikzpicture}[remember picture,overlay]
		\node[anchor=south,yshift=10pt] at (current page.south) {\fbox{\parbox{\dimexpr\textwidth-\fboxsep-\fboxrule\relax}{\copyrighttext}}};
	\end{tikzpicture}%
}

\title{\LARGE \bf
Differentiable Collision-Free Parametric Corridors}
\if\anonymous1
\author{Jon Arrizabalaga$^{1,2}$, Zachary Manchester$^{2}$, Markus Ryll$^{1}$
	\thanks{$^{1}$Munich Institute of Robotics and Machine Intelligence (MIRMI), Technical University of Munich (TUM)}
 	\thanks{$^{2}$Robotics Institute (RI), Carnegie Mellon University (CMU)}
}
\fi
\begin{document}

\maketitle
\ifcomment
    \copyrightnotice
\fi
\begin{abstract}
    This paper presents a method to compute differentiable collision-free parametric corridors. In contrast to existing solutions that decompose the obstacle-free space into multiple convex sets, the continuous corridors computed by our method are smooth and differentiable, making them compatible with existing numerical techniques for learning and optimization. To achieve this, we represent the collision-free corridors as a path-parametric off-centered ellipse with a polynomial basis. We show that the problem of maximizing the volume of such corridors is convex, and can be efficiently solved. To assess the effectiveness of the proposed method, we examine its performance in a synthetic case study and subsequently evaluate its applicability in a real-world scenario from the KITTI dataset.
\end{abstract}

\begin{flushleft}
\textbf{Code}: {\small \url{https://github.com/jonarriza96/corrgen}}\\
\textbf{Video}: {\small \url{https://youtu.be/MvC7bPodXz8}}
\end{flushleft}
\section{Introduction}\label{sec:introduction}

\noindent Path-parametric methods have gained popularity in the formulation of navigation algorithms, such as high-level planners~\cite{verschueren2016time, spedicato2017minimum, arrizabalaga2023sctomp}, reinforcement learning (RL) policies~\cite{song2021autonomous,wurman2022outracing,kaufmann2023champion} or low-level model predictive controllers (MPC)~\cite{liniger2015optimization,oelerich2024boundmpc,arrizabalaga2022towards}. The fundamental concept behind these parametric methods is to either introduce the path-parameter as an additional degree of freedom, enabling the system to  regulate its progress along the path~\cite{lam2010model, faulwasser2015nonlinear}, or to conduct a change of coordinates that project the \emph{Euclidean states} to the \emph{spatial states}, i.e., the progress along the path and the orthogonal distance to it~\cite{verscheure2009time, van2016path, arrizabalaga2022spatial}. These parametric formulations have proven successful for two primary reasons: firstly, they inherently capture the notion of advancement along the path, and secondly, spatial bounds manifest as convex constraints in the orthogonal terms of the spatial states.

Despite the increasing interest in path-parametric methods, existing collision-free space representation techniques proposed in the literature are not compatible with such formulations. The current solutions to this problem, often referred to as \emph{corridor generation}, involve decomposing the free space into a collection of convex polyhedra~\cite{deits2015computing, liu2017planning, zhong2020generating, toumieh2022voxel}. However, applying these methods to the aforementioned parametric formulations presents two significant drawbacks: Firstly, discretizing the free space into multiple convex sets introduces the \emph{polyhedron allocation problem}, wherein the assignment of a polyhedron to a particular state must be predetermined~\cite{tordesillas2019faster}. Essentially, this discretization disrupts the differentiability of the corridors with respect to the progress variable, making it challenging to integrate the collision-free corridors into optimization and learning routines. Secondly, while representing the free space with a set of hyperplanes aligns with the \emph{Euclidean} perspective, it neglects the ability to enforce convex constraints in the orthogonal components.

\begin{figure}[t]
	\centering
	\includegraphics[width=\linewidth]{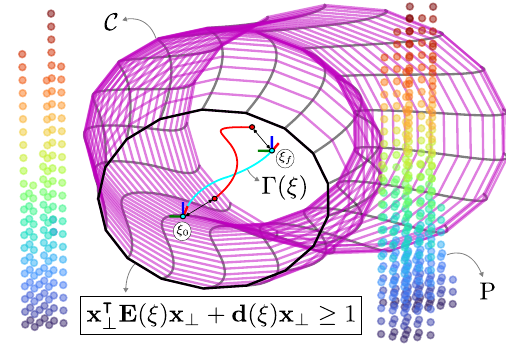}
	\caption{Given a reference path $\Gamma$ paramaterized by path-parameter $\xi$ and a point cloud $\mathrm{P}$, we present a method that computes differentiable and smooth parametric corridors $\mathcal{C}$. The reference path $\Gamma$ is depicted by the blue line with a moving frame, the colored dots refer to the point cloud $\mathrm{P}$ and the corridor $\mathcal{C}$ is given in magenta. To generate the corridor, we optimize over a parametric ellipse $\{\mathrm{E}(\xi),\mathbf{d}(\xi)\,|\,\xi\in[\xi_0,\xi_f]\}$, where besides searching for its dimensions, we also compute the offset from the reference $\Gamma$. In other words, the center of the ellipse, shown in red, is allowed to deviate from the reference $\Gamma$. Our method can run real-time and is applicable to both, planar (2D) and spatial (3D) cases.}\label{fig:explanation}
\end{figure}


In this context, where conventional collision-free generation algorithms cannot be directly applied to parametric formulations, and where obstacle avoidance primarily involves constraining the transverse states, the development of parametric collision-free corridors has been overlooked. Consequently, the corridors generated by studies utilizing parametric formulations are often tailored to specific cases and do not generalize across different scenarios~\cite{kloeser2020nmpc, arrizabalaga2022towards, romero2023weighted}. Due to the lack of a cross-compatible parametric corridor generation approach, most of the literature on path-parameterized methods prioritizes performance over safety, dropping the ability to impose spatial constraints, and consequently, leading to the loss of formal safety guarantees. In the MPC framework, this translates into an extremely fragile tuning process~\cite{romero2022model}, heavily reliant on the sophistication and patience of an expert. Similarly, in the RL scenario this entails highly penalizing excursions out of the admissible corridors, forcing the policy to learn a trade off between optimality and safety~\cite{song2023reaching, penicka2022learning}.

In light of these shortcomings, there is a need to properly address the generation of collision-free parametric corridors. Multiple research questions arise naturally, including how to parameterize the obstacle-free space in a differentiable way, and how to construct a lightweight program capable of computing the respective corridors in real-time.


In this paper, we address this need by presenting a method that, given a point cloud and a reference path, computes a collision-free parametric corridor around it. More specifically, the generated corridors are differentiable with respect to the path-parameter, making them compatible with existing optimization and learning pipelines; they can be larger than the corridors generated by existing tools; and they rely on a well-structured convex optimization formulation that can be solved in real-time. To the best of the authors' knowledge this is the first generic method that enables the computation of differentiable parametric corridors in real-time that are compatible with path-parametric formulations. To achieve this, our method relies on three main ingredients: 
\begin{enumerate}
\item We represent the obstacle-free space as a parametric off-centered ellipse with Chebyshev polynomials. 
\item We formulate a convex optimization problem -- a Linear Program (LP) for 2D planar cases and a Semidefinite Program (SDP) for 3D spatial cases -- capable of maximizing the volume of the corridor within the collision-free space.
\item We show that approximating the parametric ellipse by a diagonally-dominant matrix reduces the original SDP problem to an LP, resulting in real-time computational tractability without compromising the corridor's volumetric capacity.  
\end{enumerate}

The remainder of this paper is structured as follows: Section~\ref{sec:problem_statement} formally introduces the parametric corridor-generation problem tackled in this work, Section~\ref{sec:methodology} presents our solution by delving deeper into all three ingredients, Section~\ref{sec:experiments} shows the simulated and experimental results, and lastly, Section~\ref{sec:conclusion} presents the conclusions. 

\section{Problem statement}\label{sec:problem_statement}
\noindent In this section, we introduce the foundational concepts of our method, namely the reference path, the point cloud and the corridor. While the notation and problem description are presented in a generic spatial 3D form, we note that they also apply to the planar 2D case.

Let the occupied space be represented by a \emph{point cloud} $\mathrm{P}\in\mathbb{R}^{m\times3}$ with $m$ points
\begin{equation}\label{eq:pointcloud}
    \mathrm{P} = \left[\bm{p}_{1},\dots,\bm{p}_{m}\right]\quad \text{where}\quad\bm{p}_{1,\dots,m}\in\mathbb{R}^{3}\,,
\end{equation}
and assume $\Gamma$ to be a \emph{reference path} within the free space $\mathrm{P}\notin\Gamma$. Without loss of generalization, we define $\Gamma$ as a parametric path with a moving frame attached to it. Its respective position and orientation are given by two functions, $\bm{\gamma}\,:\,\mathbb{R}\mapsto\mathbb{R}^3$ and $\mathrm{R}_\Gamma\,:\,\mathbb{R}\mapsto\mathbb{R}^{3\times3}\subseteq\mathrm{SO}(3)$, dependant on path-parameter $\xi$:
\begin{equation}\label{eq:geom_ref}
    \Gamma = \{\xi \in[\xi_0,\xi_f] \subseteq\mathbb{R}\mapsto\bm{\gamma}(\xi) \in \mathbb{R}^3, \mathrm{R}_\Gamma(\xi) \in \mathbb{R}^{3\times3}\}\,.
\end{equation}
Additionally, we define the transverse plane $\perp_\Gamma(\xi)\in\mathbb{R}^2$ as the plane orthogonal to the path, i.e. its basis is given by the second and third column of $\mathrm{R}_\Gamma$.

The collision-free space associated to point cloud $\mathrm{P}$  around reference $\Gamma$ is denoted as a \emph{corridor} $\mathcal{C}_{\Gamma}^{\mathrm{P}}$ and encoded as an inequality on a function $\mathfrak{c}_{\Gamma}^{\mathrm{P}}\,:\,\mathbb{R}\mapsto\mathbb{R}^{c}$, such that,
\begin{align}\label{eq:corridor}
    \mathcal{C}_{\Gamma}^{\mathrm{P}} = \{\xi \in[\xi_0,\xi_f]\subseteq\mathbb{R}\,,\,\,\bm{x}_\perp\in{\perp_\Gamma(\xi)}\,\subseteq\mathbb{R}^2\mapsto\notag\\
    \mathfrak{c}_{\Gamma}^{\mathrm{P}}(\xi, \bm{x}_\perp)\leq\bm{0} \subseteq \mathbb{R}^{c}\}\,,
\end{align}
where $c$ depends on the selected free space representation, and thus, will be specified in the upcoming Section~\ref{sec:methodology}. For a visual illustration of the point cloud $\mathrm{P}$ in~\eqref{eq:pointcloud}, the reference path $\Gamma$ in~\eqref{eq:geom_ref} and the collision-free corridor $\mathcal{C}_{\Gamma}^{\mathrm{P}}$, please see Fig.~\ref{fig:explanation}. 

Having defined the preliminaries, we can now formally state the problem addressed in this paper:
\begin{problem}[\bfseries Collision-free corridor generation]\label{problem:cg}
    Given the point cloud $\mathrm{P}$ in~\eqref{eq:pointcloud} and the parametric reference path $\Gamma$ in~\eqref{eq:geom_ref}, compute a parametric corridor $\mathcal{C}_{\Gamma}^{\mathrm{P}}$, as defined in~\eqref{eq:corridor}, that is:
    \begin{itemize}
        \item[P1.1] \textbf{Collision-free:} The corridor is contained within the obstacle-free space, i.e., $\mathrm{P}\notin\mathcal{C}_{\Gamma}^{\mathrm{P}}$ .

        \item[P1.2] \textbf{Differentiable:} The corridor is (at least) two times differentiable and continuous with respect to the path-parameter, i.e., $\mathfrak{c}_{\Gamma}^{\mathrm{P}}(\xi)$ is $C^2_\xi$. 
        
        \item[P1.3] \textbf{Volume-maximizing:} The corridor encompasses the entire available space around the reference path $\Gamma$ without intersecting with the point cloud $\mathrm{P}$.
    \end{itemize}
\end{problem}

Fulfilment of P1.1 and P1.3 results from the desire of maximizing navigation performance while maintaining safety. Besides that, P1.2 guarantees the existence of continuous first and second-order derivatives, and thereby, ensures the corridor's compatibility with existing numerical methods for learning and optimization. In the remainder of this manuscript, we present a method that solves this problem. 

\begin{figure}[t]
	\centering
	\includegraphics[width=\linewidth]{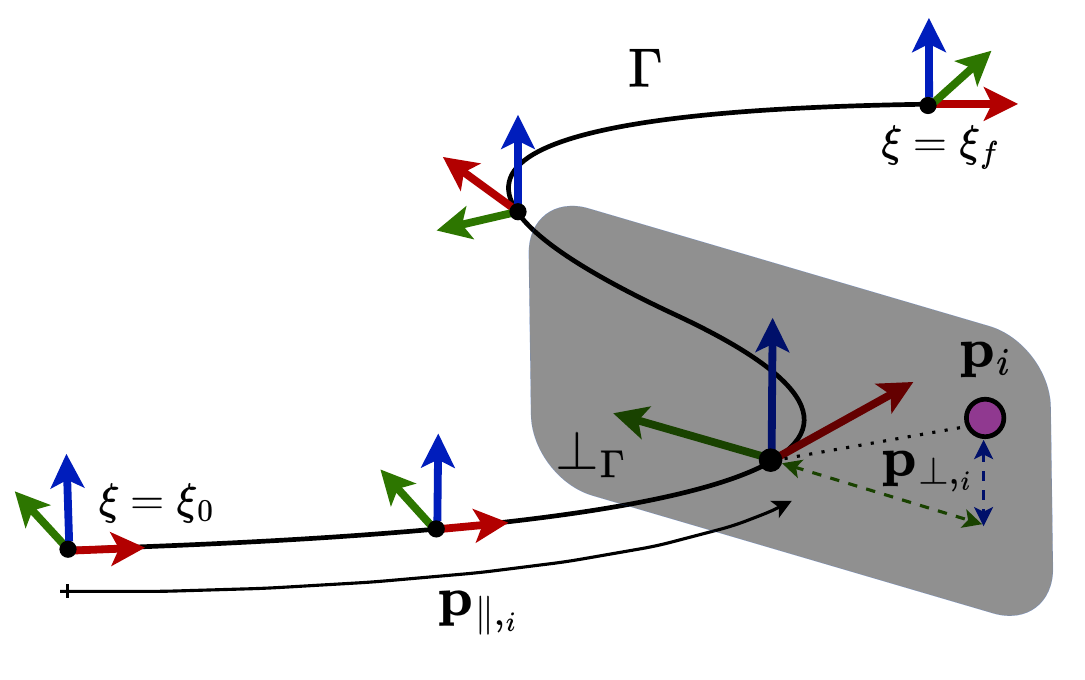}
	\caption{Projection of a point $\bm{p}_i$, represented by the pink dot, from its Euclidean coordinates onto the reference path $\Gamma$ which is parameterized by $\xi$ and has a moving frame $\mathrm{R}_\Gamma(\xi)$ attached to it. The orthogonal plane $\perp_\Gamma$ defined by the second (green) and third (blue) components of the frame is depicted in gray. The coordinates resulting from the projection are decoupled into a tangent $\bm{p}_{\parallel,i}$ and an orthogonal $\bm{p}_{\perp,i}$  component.} \label{fig:projection}
\end{figure}

\section{Solution approach}\label{sec:methodology}
\noindent The corridor-generation problem discussed in this paper is built upon (i) a projection of the point cloud $\mathrm{P}$ to the reference path $\Gamma$, (ii) a representation of the corridor as a parametric ellipse whose components are polynomials and (iii) a convex optimization problem that allows for maximizing the corridor's volume. These three elements serve as the fundamental building blocks of the proposed methodology. In the following section, we provide a more in-depth discussion of each of these components.

\subsection{Projecting the point cloud to the reference path}

\noindent Prior to selecting a specific corridor representation for \eqref{eq:corridor}, a preliminary step is necessary: projecting the point cloud from Euclidean space onto the reference path $\Gamma$, i.e., $\mathrm{P}\rightarrow\mathrm{P}_\Gamma=\left[\mathrm{P}_\parallel\,,\mathrm{P}_\perp\right]$. This process involves decomposing the point cloud into its tangential component $\mathrm{P}_\parallel$, which corresponds to the path-parameter associated with each point in the cloud, and its orthogonal component $\mathrm{P}_\perp$, representing the projection of the distance to the reference $\Gamma$ onto the transverse plane $\perp_\Gamma$. Fig.~\eqref{fig:projection} provides a visual depiction of this point projection onto the reference path.

From a mathematical standpoint, the projection is conducted in two steps. First, we find the path-parameters associated to the closest point in the reference path by
\begin{subequations}
\begin{equation}\label{eq:projection1}
    \mathrm{P}_{\parallel} = \arg \min_{\xi} || \mathrm{P} - \bm{\gamma}(\xi)||_2\,,
\end{equation}
and second, we compute the orthogonal term by projecting the distance vector to the transverse plane of the reference path:
\begin{equation}\label{eq:projection2}
    \mathrm{P}_{\perp} = \left(\mathrm{P} - \bm{\gamma}(\mathrm{P}_{\parallel})\right)\mathrm{R}_\Gamma(\mathrm{P}_{\parallel})\,.
\end{equation}
\end{subequations}
Notice that the first (tangential) term of $\mathrm{P}_\perp$ is $0$, while the remaining two decompose the distance into the second and third components of the moving frame $\mathrm{R}_\Gamma$. Besides that, the projection operation, as outlined in eqs.~\eqref{eq:projection1} and~\eqref{eq:projection2}, is fully paralellizable, and thus, it is particularly well-suited for implementation on a GPU, enabling efficient projections regardless of the size of the point cloud. 

\subsection{Selecting a parametric cross section}
\noindent Motivated by the need to formulate a corridor-generation method that is compatible with path-parameterized navigation methods and existing learning and optimization routines, we consider \emph{parametric} corridors of the form~\eqref{eq:corridor}. In other words, the corridors consist of a predefined cross section, whose dimensions evolve according to path-parameter $\xi$.

The choice of the cross section holds significant importance as it directly impacts the expressiveness and flexibility of the corridor. A cross section with fewer degrees of freedom limits the corridor's ability to accurately approximate the true free space. Hence, our objective is to select a cross section that offers maximum adaptability while ensuring differentiability\footnote{The naive way to approach this problem would be to solve a planar convex decomposition problem at the desired transverse plane. Since the number of sides associated with a polygon obtained from convex decomposition is unknown, the resulting corridor would not comply with P1.2 in Problem~\ref{problem:cg}, i.e., it would be neither differentiable nor continuous.} and computational feasibility. 

\begin{figure}[t]
	\centering
	\includegraphics[width=\linewidth]{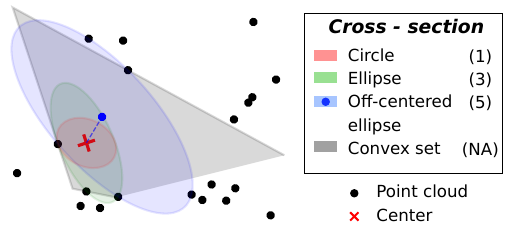}
	\caption{Area maximization for different cross section representations centered in the red cross within the transverse point cloud $\mathrm{P}_{\perp}$ represented by the black dots. The depicted cross sections are: a circle with $1$ degree of freedom (red), an ellipse with $3$ (green) and an offset-ed ellipse with a total $5$ (blue). The gray area represents the maximum convex polygon for the given point cloud encompassing the center. The offset for the off-centered ellipse is given by the blue dashed line. From this comparison, it is apparent that the higher the number the degrees of freedom, the bigger the area of the cross section.}\label{fig:cross_section}
\end{figure}

The simplest cross section consists of a circle, which offers a single degree of freedom (the radius). An ellipse adds another degree of freedom by enabling different width and height parameters. Letting the ellipse rotate further increases the decision variables to $3$. Additionally, allowing it to have an offset from the center raises the total degrees of freedom to $5$. Ultimately, by expanding the ellipse's basis with additional polynomial terms, we can create a semialgebraic set~\cite{shiota2012geometry}, and thus, provide a high degree of expressiveness to the cross section. This concept is illustrated in Fig.~\ref{fig:cross_section}, which compares the cross sections obtained by solving an area maximization problem. As can be seen, a higher number of degrees of freedom corresponds to a larger cross-sectional area.


The richness of semialgebraic sets comes at the expense of difficulties in guaranteeing the closeness and convexity of the cross section~\cite{lasserre2009convexity}. This choice makes the overall corridor generation pipeline more complicated, both from a formulation and a numerical standpoint. For this reason, we pick the next most favorable option: a rotating off-centered ellipse. Notice that, within the navigation scenario, allowing the ellipse to have an offset with respect to the reference path $\Gamma$ is of major importance, since the reference might be very close to the point cloud, specially if it is obtained from search or sampling-based methods\footnote{This also provides an alternative view on the corridor-generation problem, where corridors whose cross sections are allowed to be off-centered and contain a rich-enough representation are an appealing alternative to other hierarchical and compute-heavy mapping methods, such as Euclidean Signed Distance Functions (ESDFs)~\cite{oleynikova2017voxblox,han2019fiesta, pan2022voxfield, wang2024implicit }.}~\cite{lavalle2006planning}.

The aforementioned off-centered rotating ellipse is given by
\begin{equation}\label{eq:ellipse}
    \mathcal{E}(\mathrm{E},\bm{p_E}) = \{\bm{x}_\perp\in\mathbb{R}^{2}\,|\,(\bm{x}_\perp - \bm{p_E})^\intercal \mathrm{E} (\bm{x}_\perp - \bm{p_E})\leq{1}\}\,,
\end{equation}
where $\mathrm{E}\in S^2_{++}$ is a symmetric positive-definite matrix and $\bm{p_E}\in\mathbb{R}^2$ is the center of the ellipse. Once the cross section is defined as a rotating off-centered ellipse in~\eqref{eq:ellipse}, the next step is to path-parameterize it to sweep along the reference path from the initial point $\xi_0$ to the final point $\xi_f$. To achieve this, we employ polynomials of degree $n$ that govern the evolution of both the ellipse and its center with respect to the path-parameter $\xi$. This allows us to fully define the parametric function of the corridor introduced in~\eqref{eq:corridor} as:
\begin{align}\label{eq:corridor2}
    &\mathfrak{c}_{\Gamma}^{\mathrm{P}}(\xi,\,\bm{x}_\perp,\,\bm{c_E},\,\bm{c_P}) =\notag\\
    &\left(\bm{x}_\perp - \bm{p_E}(\xi,\bm{c_P})\right)^\intercal \mathrm{E}(\xi,\bm{c_E}) \left(\bm{x}_\perp - \bm{p_E}(\xi,\bm{c_P})\right)-1\,,
\end{align}
where $\bm{c_E}\in\mathbb{R}^{2\times2\times n+1}$ and $\bm{c_P}\in\mathbb{R}^{2\times n+1}$ are the coefficients of the polynomials associated with the ellipse matrix $\mathrm{E}$ and its centerpoint $\bm{p_E}$, respectively. From~\eqref{eq:corridor2}, it is apparent that the dimension of the inequality in~\eqref{eq:corridor} is $c=1$.

Intuitively, increasing the degree $n$ of the polynomial enhances the corridor's ability to adapt to variations along the reference path. The utilization of polynomials offers dual advantages: Firstly, it ensures that the resulting corridors exhibit inherent smoothness and differentiability with respect to the path-parameter $\xi$. Secondly, this formulation aligns with the \emph{sums of squares} framework~\cite{parrilo2000structured}, a critical aspect for formulating the volume maximization problem discussed in the subsequent subsection. 

\subsection{Corridor volume maximization}\label{subsec:volume_max}
\noindent To encapsulate the entire free space, we aim to maximize the volume of the corridor by formulating an optimization problem that finds the polynomial coefficients $\bm{c_E}$ and $\bm{c_P}$ in~\eqref{eq:corridor2}. Given that the area of the parametric ellipse $\mathcal{E}$ in~\eqref{eq:ellipse} is $A_\mathcal{E}=\pi/\sqrt{\det \mathrm{E}}\,$, the optimization problem we seek to solve is as follows:
\begin{subequations}\label{eq:ocp}
	\begin{alignat}{3}
    \max_{\bm{c_E},\bm{c_P}} V_\mathcal{E} &= \int_{\xi_0}^{\xi_f} -\det \mathrm{E}(\xi, \bm{c_E})\,d\xi\label{eq:ocp_cost}\\
	\text{s.t.}\quad &\mathrm{E}(\xi,\bm{c_E})\succ 0\,,    \quad&\xi \in \left[\xi_0,\xi_f\right]\label{eq:ocp_posdef}\\
    &\mathfrak{c}_{\Gamma}^{\mathrm{P}}(\mathrm{P}_\parallel,\,\mathrm{P}_\perp,\,\bm{c_E},\,\bm{c_P})\geq0\label{eq:ocp_Ellipse}\,,
	\end{alignat}
\end{subequations}
where~\eqref{eq:ocp_posdef} ensures that the ellipse matrix $\mathrm{E}(\cdot,\bm{c_E})$ remains positive-definite throughout the entire corridor and~\eqref{eq:ocp_Ellipse} guarantees that all the points in the cloud lie outside the corridor.

The current form of the optimization problem~\eqref{eq:ocp} is defined in the continuous domain, rendering it infeasible for direct implementation on a computer. Moreover, it is characterized by nonlinearity and nonconvexity. To make it computationally tractable, we reformulate the problem and introduce several modifications aimed at convexifying it. In particular, we address four key modifications:

Firstly, we discretize the continuous parts of the optimization problem, namely the cost function~\eqref{eq:ocp_cost} and the positive-definite constraint~\eqref{eq:ocp_posdef}, into $N$ evaluations. This discretization is valid when $N$ is significantly larger than the order of the polynomials, i.e., $N \gg n$. Secondly, to avoid the nonlinearities associated with the determinant of the ellipse in~\eqref{eq:ocp_cost}, we approximate it with the trace of the matrix, thus yielding $\det\mathrm{E} \approx \text{tr}(\mathrm{E})$. Thirdly, we prevent unbounded growth in obstacle-free areas by introducing upper bounds on the dimensions of the parametric ellipse. To this end, we modify the point cloud $\mathrm{P}$ by introducing a wrapper around the path $\Gamma$, which not only sets an upper limit but also allows us to exclude points located further away than the wrapper itself, thereby reducing the total number of constraints in the optimization problem\footnote{An alternative method is to decouple constraint~\eqref{eq:ocp_posdef} into $\mathrm{E} - \ubar{\lambda}\mathrm{I} \succ 0$ and $-\mathrm{E} + \bar{\lambda}\mathrm{I} \succ 0$, where $\ubar{\lambda}$ and $\bar{\lambda}$ represent the maximum and minimum eigenvalues of the ellipse matrix $\mathrm{E}\,$, respectively. These eigenvalues correspond to the maximum $\bar{l}$ and minimum $\ubar{l}$ axis sizes of the ellipse, related by $\ubar{\lambda} = 2/\sqrt{\bar{l}}$ and $\bar{\lambda} = 2/\sqrt{\ubar{l}}$.}. As a consequence, the optimization problem is always bounded and feasible. The reduced point cloud, obtained after introducing the wrapper and removing the points outside of it, is denoted as $_{W}\mathrm{P}\in\mathbb{R}^{w\times3}$, where $w$ is the number of remaining points. Fourth, to address the nonlinearities in~\eqref{eq:ocp_Ellipse}, we reformulate the corridor's parametric definition as
\begin{align}\label{eq:corridor3}
    \mathfrak{d}_{\Gamma}^{\mathrm{P}}(\xi,\,&\bm{x}_\perp,\,\bm{c_E},\,\bm{c_D}) =\notag\\
    &\bm{x}_\perp^\intercal \mathrm{E}(\xi,\bm{c_E}) \bm{x}_\perp + \bm{d}(\xi,\bm{c_D})\bm{x}_\perp-1\,,
\end{align}
where $\bm{d}(\xi,\bm{c_D})\in\mathbb{R}^2$ is a polynomial on $\xi$ with coefficients $\bm{c_D}\in\mathbb{R}^{2\times n+1}$. This term allows for the aforementioned offset from the reference path $\Gamma$, while removing the nonlinearities of~\eqref{eq:corridor2}.

By incorporating all four modifications, the approximation to the original problem~\eqref{eq:ocp} that we actually solve is given by:
\begin{subequations}\label{eq:sdp}
	\begin{alignat}{3}
    \min_{\bm{c_E},\,\bm{c_D}} -V_\mathcal{E} &\approx \sum_{i=1}^{N} \text{tr}(\mathrm{E}\left(\xi_i,\bm{c_E})\right)\label{eq:sdp_cost}\\
	 \text{s.t.}\quad &\mathrm{E}(\xi_i,\bm{c_E})\succ 0\,, \quad\quad i = 1,\dots,N\label{eq:sdp_constr1}\\\
    &\mathfrak{d}_{\Gamma}^{\mathrm{P}}(_{W}\mathrm{P}_\parallel,\,_{W}\mathrm{P}_\perp,\,\bm{c_E},\,\bm{c_D})\label{eq:sdp_constr3}\geq0\,.
	\end{alignat}
\end{subequations}
In its present state, \eqref{eq:sdp} is a convex optimization problem  with $5n$ decision variables and $N\,w$ constraints. More specifically, it is a Semidefinite-Program (SDP)~\cite{vandenberghe1996semidefinite}, a type of problem that can be efficiently solved with off-the shelf open-source solvers~\cite{yamashita2003implementation,andersen2020cvxopt,clarabel,odonoghue:21,garstka2021cosmo}. It is worth highlighting that, for high polynomial degrees $n$, we avoid the problem becoming ill-conditioned by choosing a Chebyshev polynomial basis~\cite{mason2002chebyshev}.

Since the positive-definite constraint in~\eqref{eq:sdp_constr1} is a two-dimensional Linear Matrix Inequality (LMI), further relaxations allow for approximating SDP~\eqref{eq:sdp} into a more restricted and lightweight formulation, such as a Second-Order-Cone-Program (SOCP) or a Linear-Program (LP) ~\cite{ahmadi2019dsos}. Aiming to maximize computational efficiency, we choose the latter option, which replaces the positive-definite constraint on $\mathrm{E}$ with diagonal dominance, i.e., both diagonal terms $\mathrm{E}_{11}$ and $\mathrm{E}_{22}$ need to be bigger than the off-diagonal one $\mathrm{E}_{12}$. Given that all diagonally dominant matrices are positive-definite ~\cite{ahmadi2019dsos}, we replace constraint~\eqref{eq:sdp_constr1} by $\mathrm{E}_{11}>\mathrm{E}_{12}$ and $\mathrm{E}_{22}>\mathrm{E}_{12}$, and thus, reduce the original SDP problem~\eqref{eq:sdp} into an LP.

 \begin{figure*}[t]
\centering
\includegraphics[width=\textwidth]{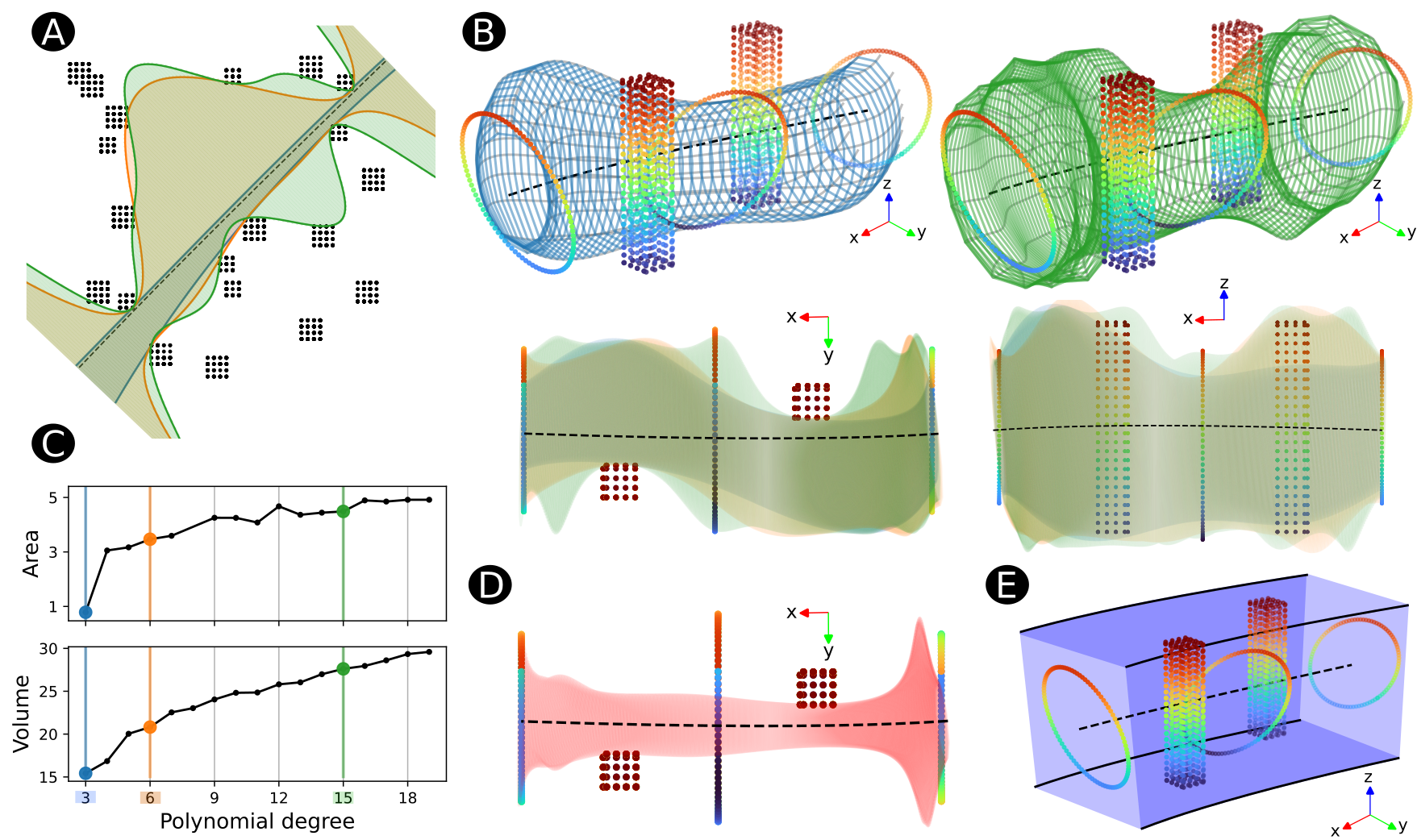}
\caption{2D and 3D corridors generated by our method for different polynomial degrees: $3$ in blue, $6$ in orange, and $15$ in green. In the planar case (A)  the point cloud is depicted in black, while in the spatial case (B) it is colored according to the height. The black dashed line represents the reference path $\Gamma$. The correlation between the corridor's area -- applicable to (A) -- and volume -- applicable to (B) -- with respect to the polynomial degree is plotted in (C). As the degree increases, the respective parametric areas and volumes increase. This trend is confirmed in (A) and (B), where the expressiveness of the sampled corridors augments with the degree. Panel (D) shows the corridor associated with a parametric ellipse centered on reference $\Gamma$, making apparent the benefits of optimizing over the offset. Panel (E) depicts the wrapper that prevents the 3D corridors in (B) and (D) from becoming unbounded.}\label{fig:simulated_results}
\end{figure*}
\subsection{The planar 2D case}
\noindent The computation of planar corridors is a subset of the problem introduced in the previous subsection. In this case, the corridor is defined by the area enclosed between two polynomials $\bm{b_+}(\xi,\bm{c_{B_+}})\in\mathbb{R}$ and $\bm{b_-}(\xi,\bm{c_{B_-}})\in\mathbb{R}$, situated on the positive and negative sides of the orthogonal plane $\perp_\Gamma$. Here, $\bm{c_{B_+}}$ and $\bm{c_{B_-}}$ represent the corresponding coefficients. The optimization problem seeks to maximize the area, which involves tailoring problem~\eqref{eq:sdp} as follows:
\begin{subequations}\label{eq:lp2d}
	\begin{alignat}{3}
    \min_{\bm{c_{B_+}},\, \bm{c_{B_-}}} \hspace{-2mm}-A_{\bm{b}} &\approx -\sum_{i=1}^{N} \bm{b}_+\left(\xi_i,\bm{c_{B_+}}\right) + \bm{b}_-\left(\xi_i,\bm{c_{B_-}}\right)\label{eq:lp2d_cost}\\
	\text{s.t.}\quad &\bm{b}_+\left(\xi_i,\bm{c_{B_+}}\right)>0\,,    \quad i = 1,\dots,N\label{eq:lp2d_constr1}\\
 &\bm{b}_-\left(\xi_i,\bm{c_{B_-}}\right)<0\,,    \quad i = 1,\dots,N\label{eq:lp2d_constr2}\\
 &\bm{b}_+\left(_{W}\mathrm{P}^+_\parallel\,,\,\bm{c_{B_+}}\right) \leq \,_{W}\mathrm{P}^+_\perp\,,\label{eq:lp2d_constr3}\\
 &\bm{b}_-\left(_{W}\mathrm{P}^-_\parallel\,,\,\bm{c_{B_-}}\right) \leq \,_{W}\mathrm{P}^-_\perp\,,\label{eq:lp2d_constr4}
	\end{alignat}
\end{subequations}
where $_{W}\mathrm{P}^{+}$ and $_{W}\mathrm{P}^{-}$ separate the point cloud in the positive and negative sides of the orthogonal plane $\perp_\Gamma$. The resulting problem~\eqref{eq:lp2d} is an LP with $2n$ decision variables and $2N\,w$ constraints.
\section{Experiments}\label{sec:experiments}
\noindent To evaluate our approach, we split the experimental analysis into two parts: First, we get a better understanding of the method's performance by focusing on two illustrative synthetic case studies. Second, we analyze its behavior in a real-world scenario obtained from the KITTI Vision Benchmark Suite~\cite{Geiger2013IJRR}. 
\subsection*{Numerical implementation}
\noindent In all evaluations, we compute the reference $\Gamma$ by first determining a collision-free path using kinodynamic path-planning~\cite{webb2013kinodynamic} and then parameterizing it with an interpolation algorithm based on Pythagorean hodograph curves~\cite{vsir2007}. It is important to note that these choices are entirely modular and independent of the presented corridor-generation method.  When benchmarking against convex decomposition, we compare against~\cite{liu2017planning},  the implementation of which is available online\footnote{\label{footnote:decomp}\url{https://github.com/sikang/DecompUtil}}. Other than that, we discretize the problem into $N=100$ evaluations and use Clarabel~\cite{clarabel} and ProxQP~\cite{bambade2023proxqp} to solve the underlying SDPs and LPs, respectively.
 \begin{figure*}[t]
\centering
\includegraphics[width=\textwidth]{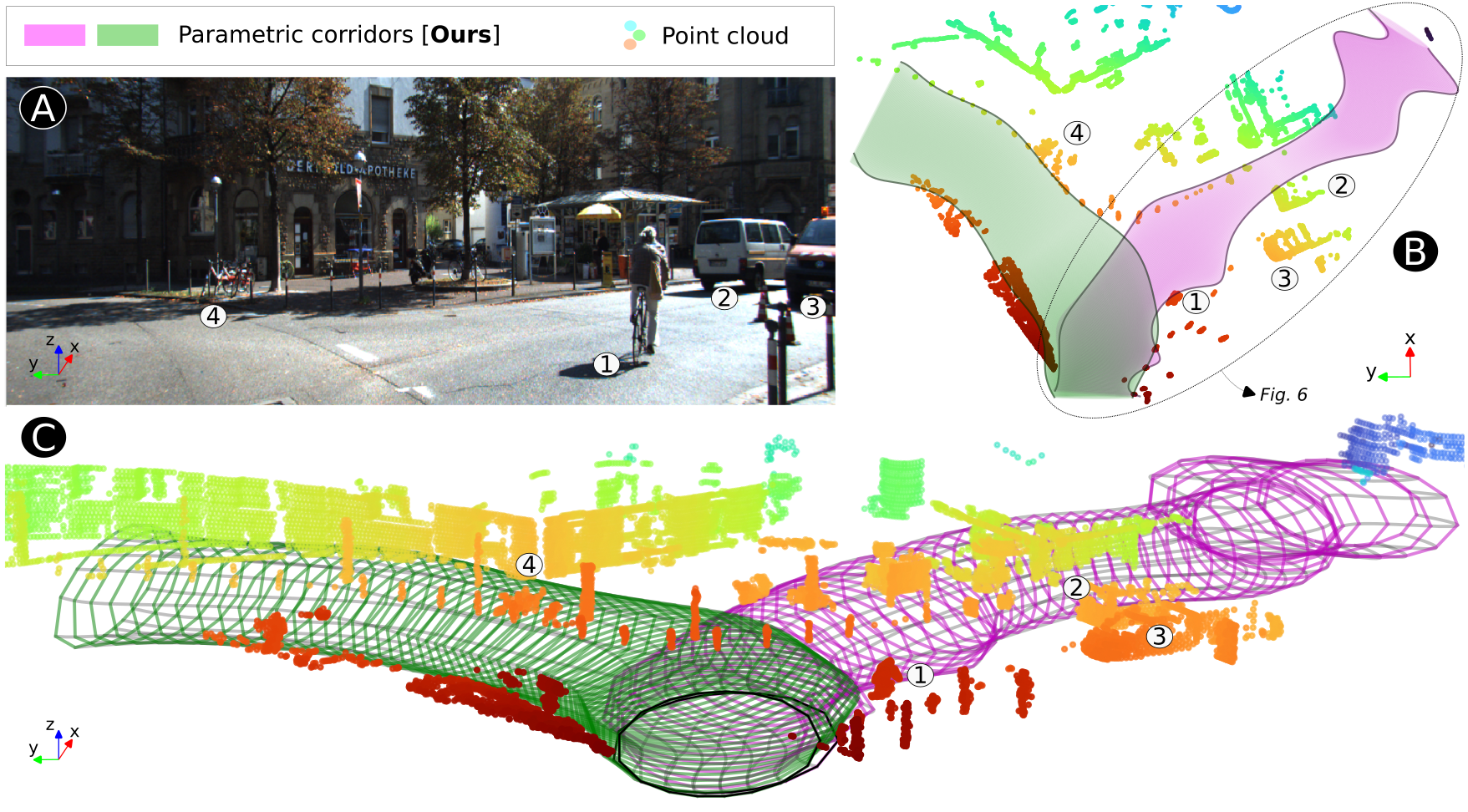}
\caption{Two different 3D corridors for polynomial degree $n=9$ generated by our method in a real-world challenging driving scenario, where the road bifurcates into two narrow lanes, with other cars and cyclists. The case study is part of the KITTI Vision Benchmark Suite~\cite{Geiger2013IJRR}. Panel (A) shows an RGB camera-based overview of the scene, while the respective point cloud and corridors are given in panels (B) --top view-- and (C) --isometric view--.  The point cloud is color-coded based on proximity to the camera's location. To facilitate the comprehension of the environment, numbered markers in white (from 1 to 4), associated with relevant features in the scene, have been added to all panels. For a detailed analysis of the corridor circled by the dashed line in Panel (B), please refer to Fig.~\ref{fig:experiment_results_2}.\label{fig:experiment_results_1}}
\vspace{-2mm}
\end{figure*}
\subsection{An illustrative toy-example}
\noindent To better comprehend our corridor-generation method, we examine its performance on two different 2D and 3D exemplary case studies created by randomly positioning a series of columns and rings. In particular, we compare the corridors obtained for various polynomial degrees $n$ and we showcase the importance of optimizing over the offset with respect to reference path $\Gamma$.

The resulting solutions are illustrated in Fig.~\ref{fig:simulated_results}. In panels (A) and (B), the corridors obtained for degrees $3$, $6$, and $15$ are presented in blue, orange, and green, respectively, for both 2D and 3D cases. From these corridors, it becomes evident that increasing the degree $n$ enhances the corridor's expressiveness: the blue corridor is the smallest, while its capacity to adapt to the environment increases with the orange, resulting in the largest corridor for green. In panel (C), we validate this observation by demonstrating the correlation between area/volume and polynomial degree for multiple degrees. Panel (D) shows the top-down view of the corridor obtained from a centered (no offset) parametric ellipse, while panel (E) depicts the wrapper used to bound the computation of the three-dimensional corridors in (B) and (D).

The analysis of our results yields two significant conclusions: Firstly, we observe that increasing the polynomial degree leads to an expansion in the area/volume of the corridor. This augmentation is particularly notable for lower degrees; however, it becomes less pronounced for higher degrees. Secondly, the ability to optimize over the ellipse's center point emerges as a crucial factor. As illustrated in panel (D) of Fig.~\ref{fig:simulated_results}, constraining the ellipse to be centered on the reference path $\Gamma$ limits the achievable maximum volume, which is specially detrimental when the reference path is close to obstacles. However, allowing for deviations in the center point makes the corridor generation method agnostic to the underlying reference path.


 \begin{figure*}[t]
\centering
\includegraphics[width=\textwidth]{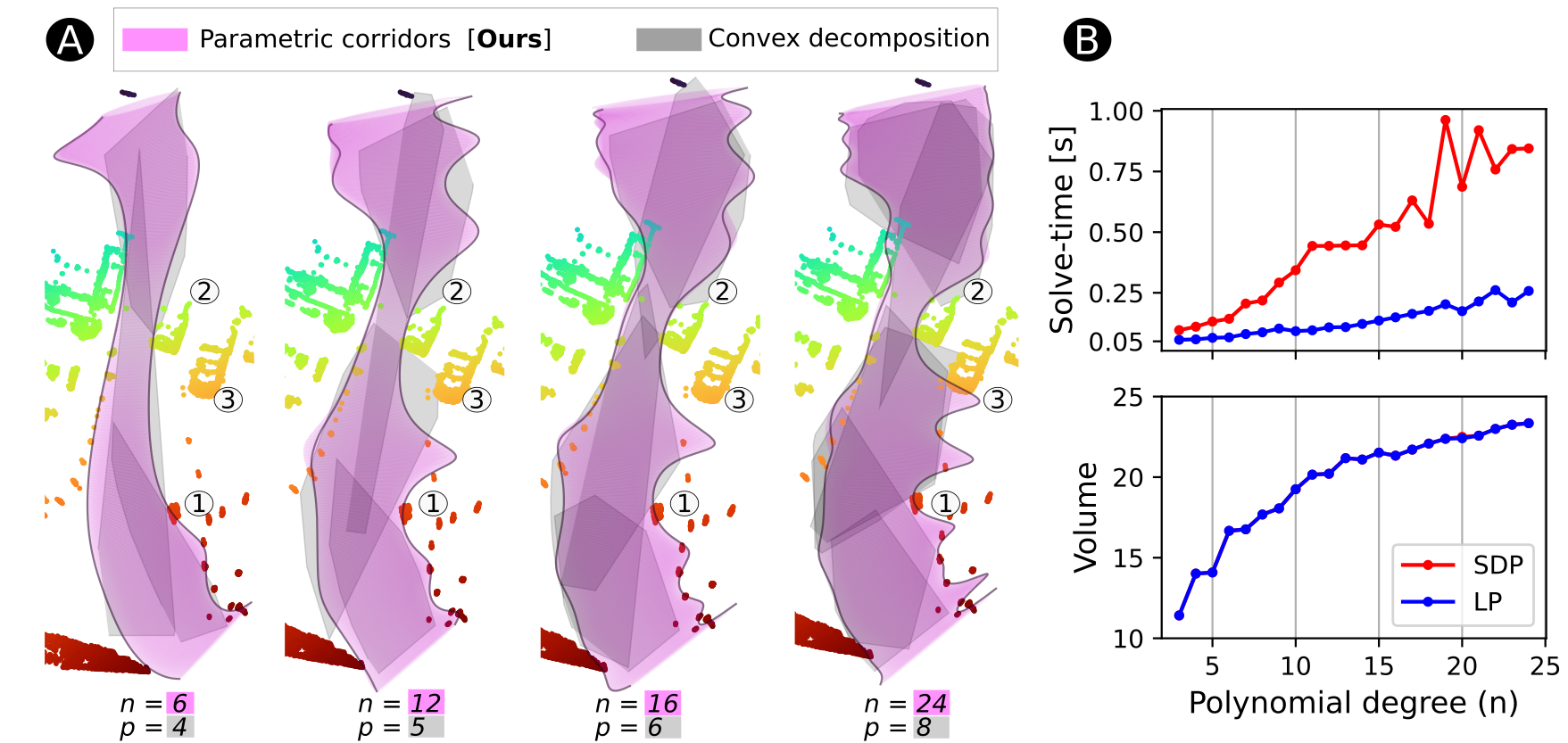}
\caption{An in-depth study of the real-world scenario depicted in Fig.~\ref{fig:experiment_results_1}. Panel (A) illustrates the top view of corridors obtained by progressively increasing the polynomial degree $n$, alongside the corridors resulting from decomposing the free space into $p$ convex sets, depicted in gray. Panel (B) shows the evolution of corridor's volume and computation times with respect to the polynomial degree for both SDP and LP problems.\label{fig:experiment_results_2}} 
\vspace{-5mm}
\end{figure*}
\subsection{Real-world case study}
\noindent Having gained a deeper understanding on the corridor generation method, we proceed to assess its applicability to a real-world scenario taken from the KITTI dataset~\cite{Geiger2013IJRR}. In particular, we focus on a roundabout intersection where a bifurcation into two lanes allows for the computation of two corridors. This intersection presents numerous challenges due to its various features, including cyclists, cars, and narrow roads. 

The resulting corridors, along with the lidar-based point cloud and a raw RGB color camera view, are shown in Fig.~\ref{fig:experiment_results_1}. The numbers added to the image in panel (A) and the point clouds in panels (B) and (C) help to understand the landscape and locate the aforementioned features. From these visualizations it is apparent that the computed corridors encapsulate the free space, extending across the entire width of the road, and adeptly accommodating other vehicles, bicycles, and road boundaries.

To gain insight into how our method compares to the well-established convex decomposition~\cite{liu2017planning}, we refer to Fig.~\ref{fig:experiment_results_2}-A, where the top views of the corridors obtained from sequentially increasing the polynomial degree $n$ are depicted alongside the corridors resulting from decomposing the free space into $p$ convex sets. The results show that both methods encompass very similar spaces in all four cases. However, the hyperparameters differ significantly. In convex decomposition, the quantity and distribution of polyhedra are determined based on the number of segments within a precomputed linear path, which may compromise the corridor's volume if the reference path has few segments. In contrast, our method employs the polynomial degree $n$ as its unique hyperparameter, while remaining agnostic to the underlying reference. Additionally, Fig.~\ref{fig:experiment_results_2}-A also depicts how a parametric cross section with a polynomial basis result in a continuous and smooth space representation, contrasting with the discreteness intrinsic to convex decomposition.

Besides the comparison with convex decomposition, Fig.~\ref{fig:experiment_results_2}-A  further confirms the tendency observed in the previous synthetic case studies, namely that increasing the polynomial degree results in larger corridors. In particular, the corridor associated with $n=24$ shows to encompass the entirety of the available space.

To reflect on the trade off between the volume gains obtained from increasing the polynomial degree $n$ and the corresponding computational burden, we refer to Fig.~\ref{fig:experiment_results_2}-B. Here, we show the volumes and solve-times obtained from computing the pink corridor in Fig.~\ref{fig:experiment_results_1} for various polynomial degrees, ranging from $n=3$ up to $n=25\,$, for both the original SDP and the approximated LP problems. From these results our conclusions are twofold: First, similar to the observations in the previous case studies, the increase in volume is significant for lower degrees and it becomes marginal for higher ones. Second, the approximated LP yields exactly the same corridor volumes as the original SDP, while showing major computational speedups.

We attribute this to two reasons: On the one hand, the cost function~\eqref{eq:sdp_cost} maximizes the trace, practically implying that the matrix of the resulting corridors are diagonally dominant. On the other hand, the freedom to optimize over the offset with respect to the reference $\Gamma$ overcomes the limitations inherited from requiring matrix $\mathrm{E}$ to be diagonally dominant, resulting in corridors capable of encapsulating the available space, while maintaining diagonally dominant matrices.

These results suggest that our method is capable of computing differentiable, continuous and smooth parametric corridors at approximately 5 to 20 Hz, rendering the suggested method real-time deployable.





\section{Conclusions}\label{sec:conclusion}
\noindent In this work, we proposed a corridor-generation method that allows the collision-free space to be represented in a continuous and differentiable manner. For this purpose, we predefine the cross section of the corridor as an off-centered ellipse whose components are parameterized by Chebyshev polynomials. Seeking a complete representation of the obstacle-free space, we show that the corridor volume maximization problem can be formulated as a convex optimization problem over the polynomial's coefficients. More specifically, the problem is a Linear Program (LP) in the 2D planar case and a Semidefinite Program (SDP) in the 3D spatial case. For the latter, we show that approximating the parametric ellipse by a diagonally dominant matrix also reduces it to an LP, resulting in significant computational speed improvements without compromising the corridor's volumetric capacity.

The performance of our method has been evaluated in synthetic case studies and further validated in a real-world scenario from the KITTI dataset. These results demonstrate that the corridors generated by our method achieve comparable volumetric capacity to those generated by the discrete convex decomposition method, eliminating the requirement to predefine the number of convex sets. Finally, the continuous, smooth, and differentiable nature of the obtained corridors, coupled with their real-time deployability, underscores the method's suitability for optimization and learning-based path-parametric navigation algorithms.

\newcommand{\BIBdecl}{\setlength{\itemsep}{0.09 em}} 
\bibliographystyle{IEEEtran}
\bibliography{Differentiable_Parametric_Corridors}

\end{document}